\documentclass[]{ceurart}

\usepackage{listings}
\usepackage{amsmath}
\usepackage{graphicx}
\usepackage{booktabs}
\begin{document}

\copyrightyear{2026}
\copyrightclause{Copyright for this paper by its authors.
  Use permitted under Creative Commons License Attribution 4.0
  International (CC BY 4.0).}

\conference{Pre-print Ital-IA 2026: 6th National Conference on Artificial Intelligence,
organized by CINI, June 18--19, 2026, Rome, Italy}

\title{Divide, Deliberate, Decide: A Multi-Agent Framework for Fine-Grained Egocentric Action Recognition}

\author[1]{Alessandro Sottovia}[%
  orcid=0009-0004-8870-8261,
  email=asottovia@unibz.it,
]

\cormark[1]

\author[1]{Alessandro Torcinovich}[%
  orcid=0000-0001-8110-1791,
  email=Alessandro.Torcinovich@unibz.it,
]

\author[1]{Oswald Lanz}[%
  orcid=0000-0003-4793-4276,
  email=Oswald.Lanz@unibz.it,
]

\address[1]{Faculty of Engineering, Free University of Bozen-Bolzano, Bolzano, Italy}

\cortext[1]{Corresponding author.}

\begin{abstract}
Fine-grained action recognition in egocentric video is challenging for Vision-Language Models (VLMs): actions often differ only in small visual cues, and a single model tends to be biased toward a subset of these cues. We propose \emph{Divide, Deliberate, Decide}, a fully-local, zero-shot multi-agent framework in which (i) a VLM \emph{orchestrator} chunks the video and proposes a top-$k$ candidate label list per segment, (ii) an ensemble of \emph{heterogeneous} VLM specialists, drawn from different open model families, engages in a structured deliberation that includes a peer-consultation round of questions, and (iii) agent rankings are aggregated with a Borda count and the orchestrator re-ranks its own prediction in light of the specialists' evidence. The entire pipeline runs locally with no fine-tuning. Experiments show that our method positively improves zero-shot action recognition performance over the baseline, highlighting the influence of a heterogeneous deliberation step, showing that the gain stems from decorrelated model priors rather than from additional compute.
\end{abstract}


\maketitle

\section{Introduction}

Recognizing fine-grained actions in egocentric video is central to intelligent systems deployed in industrial assistance, AR-guided training, and human robot collaboration scenarios. Unlike coarse activity classification, fine-grained recognition requires distinguishing actions that are visually almost identical for example, \texttt{"take a screw"} versus \texttt{"put a screw"}, where the correct label differs by a small spatial or temporal cue such as whether a tool is held, which object is in contact, or the direction of hand motion. Vision-Language Models (VLMs) are strong at holistic video description but display well-documented weaknesses in fine-grained temporal reasoning. A model of moderate size, queried on a short video clip, tends to anchor on the most salient visual token (typically the dominant object) and miss the discriminative cue \cite{cai2024temporalbench, li2024mvbench}. Scaling the model has been proved as one remedy \cite{wang2025inference}, but it is neither always possible nor always desirable: deployment on-premise or on-edge in industrial settings often rules out large proprietary VLMs for latency, privacy, or cost reasons.

As an alternative, in line with recent orchestration-based LLM paradigms \cite{Taulli_2025_autogen,liang2024encouraging}, we propose to refine a classifier's predictions, with those coming from an ensemble of small VLM agents. In particular, the ensemble \emph{deliberates} on such predictions, i.e., it simulates a Q\&A session to update the agents priors, thus converging to a shared conclusion. The heterogeneity of the ensemble matters: models trained on different data mixtures and with different visual backbones fail in different ways, and their disagreements are informative. The challenge is to structure the interaction so that agents can compare evidence and revise their positions without collapsing into trivial consensus.  

We present \emph{Divide, Deliberate, Decide}, a fully-local agentic framework organized in three stages: \textit{(i) Divide:} An \emph{orchestrator} model (i.e., a VLM) divides the input video into chunks, proposes segment boundaries, and produces a preliminary top-$k$ action list per segment, \textit{(ii) Deliberate:} An ensemble of heterogeneous \emph{specialist} agents (i.e., smaller VLMs) engage in a structured deliberation per segment to present their own action rankings. \textit{(iii) Decide.}  Agent rankings are combined via Borda count \cite{van2000variants} -- i.e., a rank-aggregation rule rewarding candidates ranked consistently high -- and the orchestrator updates its initial prediction based on the Borda winner and the individual agent votes.

Our central empirical questions are: (Q1) Does a VLM orchestrator benefit from structured multi-agent deliberation improving consistently its performance? (Q2) How the deliberation affects the orchestrator's final predictions? (Q3) Is heterogeneous deliberation -- with agents coming from different model families -- preferred over homogeneous? (Q4) Does a VLM orchestrator, in zero-shot, produce temporal segment boundaries of sufficient quality?
\section{Related Work}

\textbf{Fine-grained egocentric action recognition.} Egocentric benchmarks such as MECCANO \cite{Ragusa_2023_Meccano}, EPIC-KITCHENS \cite{Damen_2022_EPIC_KITCHENS}, and Ego4D \cite{Grauman_2025} have exposed the gap between holistic activity understanding and fine-grained action labelling. Methods based on dedicated action-recognition backbones handle short clips well but require per-dataset supervision and do not transfer naturally to open-vocabulary prompting.

\textbf{Multi-agent LLM systems and deliberation.}
The ReAct paradigm \cite{yao2022react} and multi-agent frameworks such as AutoGen \cite{Taulli_2025_autogen} and CAMEL \cite{li2023camel} demonstrate that decomposing reasoning across specialized agents improves performance on tasks that a single model handles unreliably. Self-consistency \cite{wang2022self} and self-refinement have shown that soliciting multiple candidate answers and re-ranking improves LLM accuracy. Our work transfers these ideas to the video domain but departs from prior work in two ways: our agents are \emph{heterogeneous} (different model families, not multiple roles of the same model), and the deliberation protocol is explicitly structured around a peer-consultation round that lets agents probe each other's visual evidence.   


\section{Method}

\begin{figure}
  \centering
  \includegraphics[width=\linewidth]{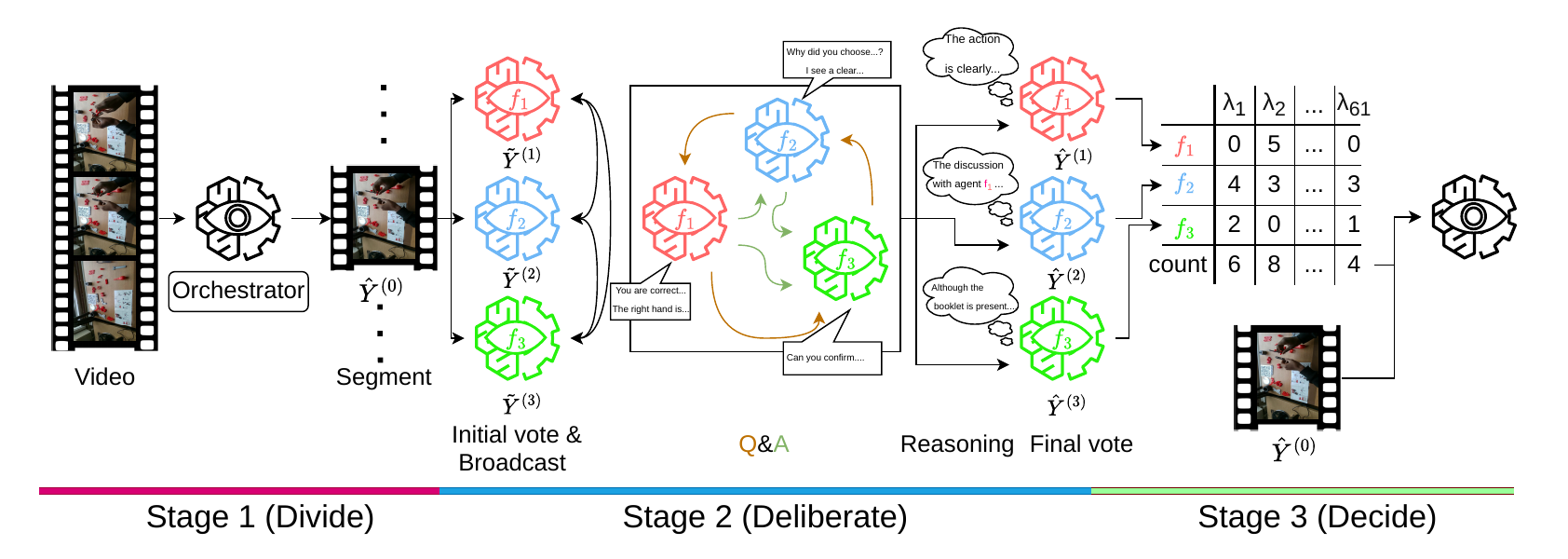}
  \caption{Three-stage pipeline: orchestrator segmentation (Stage~1),
  six-turn heterogeneous-agent deliberation (Stage~2), and
  aggregation with orchestrator re-ranking (Stage~3).}
  \label{fig:architecture}
\end{figure}

Action recognition aims at identifying the actions performed in a video, by assigning an action label to each of its frames. In our setting, we are given a video $V \in \mathbb{R}^{C \times W \times H \times T}$ and a set of action labels $\Lambda = \{\lambda_1, \ldots, \lambda_m\}$. We adopt a three-stage pipeline described in Figure~\ref{fig:architecture}.

In stage 1 (Divide), the orchestrator performs an initial segmentation of the video and performs an initial top-$k$ ranking of the actions present in each segment. In stage 2, the specialists deliberate over each segment and the initial prediction, generating their own rankings. In stage 3, the specialists' rankings are aggregated via Borda count, and the orchestrator is asked to re-rank the segments in light of the new evidence. In the following, we describe the three stages in detail.

\subsection{Stage 1: Divide}
Often a VLM cannot process an entire video in a single pass, due to excessive length. Therefore, we first split $V$ into chunks $V_1, \ldots, V_C$ of maximum length $t_{\text{max}}$. Additionally, each chunk $V_c$ is temporally subsampled at a lower frame rate, and its frames are resized to a uniform resolution. The orchestrator is then prompted with the chunk together with the labels $\Lambda$ and is tasked to divide the chunk in a list of candidate segments $s_1, \ldots, s_{v_{c}}$, each containing a single action. For each segment, the orchestrator also generates an initial candidate list of the $k$ most fitting actions $\hat{Y}^{(0)} = \{\hat{y}_1^{(0)}, \ldots, \hat{y}_k^{(0)}\}$, describing that segment.\looseness=-1

\subsection{Stage 2: Deliberate}
\label{sec:Stage2}

In the second stage, a set of VLM specialists $f_i$, $i = 1, \ldots, p$ are tasked to discuss about the initial ranking provided by the orchestrator. The deliberation process is so structured. First, each agent receives a segment and its associated ranking $\hat{Y}^{(0)}$, and outputs its own top-$k$ \emph{prior} candidate action list $\tilde{Y}^{(i)}$. The priors are then shared among all agents. When they disagree, agents can ask each other factual questions to probe the disputed visual cue (e.g., \texttt{"Given that [...], is action A or B the more accurate description of the action?" "The action is best described as [...]"}). To limit the computational overhead, we limit each agent to ask one question only. Finally, all agents receive the full exchange and update their rankings, obtaining the posteriors $\hat{Y}^{(i)}$. This protocol is deliberately designed so that agents can see where they disagree, probe the specific visual cue underlying the disagreement, and revise without external pressure toward consensus. Unanimity is not enforced, agents may retain their original ranking if peer responses do not steer them.

\subsection{Stage 3: Decide}

At the end of the deliberation, we aggregate the rankings over the specialists. Given a segment, for each of its candidate actions $y \in \bigcup_{i = 1}^p \hat{Y}^{(i)}$, we compute a Borda score \cite{van2000variants}:
\begin{equation*}
  B(y) = \sum_{i = 1}^p w\left(r_i(y)\right),
\end{equation*}
where $r_i(y): \Lambda \rightarrow \mathbb{N}$ is the rank of $y$ in agent $f_i$'s posterior ranking (the lower, the better. $k + 1$, if $y$ is not present), and $w(r) = k - (r - 1)$ is a discrete linear weighting scheme. The Borda-scores entail the specialists' aggregated top-$k$ ranking $\hat{Y}^{(B)}$.

$\hat{Y}^{(0)}, \ldots, \hat{Y}^{(p)}$, $\hat{Y}^{(B)}$ and its Borda scores are then passed as input to the orchestrator, which is prompted to produce a \emph{final} ranking $\hat{Y}^{(F)}$. Critically, the orchestrator can only select from its own and the agents' predictions, and it is not allowed to introduce new labels at this stage. This constraint ensures that the agent deliberation is the only new source of evidence informing the final decision.

\section{Experimental Setup}
\label{sec:setup}


We evaluate on the MECCANO dataset~\cite{Ragusa_2023_Meccano}, an egocentric benchmark recorded at 12\,fps of a person assembling a Meccano construction set. The annotation vocabulary contains 61 fine-grained action classes composed of \{object, motion, tool\} triples (e.g.\ \texttt{take\_red\_perforated\_bar}). For each action segment we temporally sub-sample to $\approx 8$ frames ($\approx40\%$ of the original 12\,fps), in line with prior work on egocentric VLM  that reports this regime as the sweet spot between temporal coverage and context length~\cite{bianchi2026skillformer, bianchi2026profvlm}. All models run locally through Ollama on a single workstation with one NVIDIA RTX~6000~\cite{NVIDIA_RTX6000}, no fine-tuning is performed and all inference is zero-shot. The orchestrator is Qwen3.5-27B, the three specialists are \texttt{qwen3.5:9b}, \texttt{ministral-3:8b} and \texttt{gemma4:e4b}. The full framework takes on average, $\approx$ 5 minutes per segment, while the orchestrator (baseline) alone requires, $\approx$ 40 seconds.

\begin{table}
  \caption{Top-1 / top-5 accuracy (\%) on the MECCANO test split.}
  \label{tab:main}
  \centering
  \setlength{\tabcolsep}{4pt}
\begin{tabular}{l rr rr rr rr}
    \toprule
    & \multicolumn{2}{c}{\textbf{Baseline}}
    & \multicolumn{2}{c}{\textbf{Ours}}
    & \multicolumn{2}{c}{\textbf{B. (GTB)}}
    & \multicolumn{2}{c}{\textbf{Ours (GTB)}} \\
    \cmidrule(lr){2-3}\cmidrule(rr){4-5}\cmidrule(lr){6-7}\cmidrule(lr){8-9}
    \textbf{Video} & Top-1  & Top-5 & Top-1 &  Top-5 & Top-1 & Top-5 & Top-1 & Top-5 \\
    \midrule
    0008 & 13.0 & 32.8 & 16.4 & 47.1 &  19.3 & 36.7 & 25.7 & 37.9  \\
    0009 &  9.9 & 18.8 & 12.9 & 40.6 &  18.0 & 38.4 & 24.2 & 38.8\\
    0010 & 23.4 & 38.5 & 24.0 & 55.7 &  18.8 & 38.5 & 23.7 & 40.5\\
    0011 & 11.0 & 35.6 & 14.7 & 45.5 &  20.5 & 41.0 & 22.5 & 44.2 \\
    0012 &  9.6 & 24.4 & 11.1 & 36.3 &  13.0 & 31.8 & 14.6 & 35.4 \\
    0019 & 11.9 & 23.9 & 20.1 & 51.6 &  18.5 & 39.5 & 18.5 & 40.8  \\
    0020 & 15.8 & 28.2 & 18.1 & 37.9 &  14.7  & 32.3 & 16.3 & 32.9\\
    \midrule
    \textbf{Avg} & 13.5 & 28.9 & \textbf{16.8} & \textbf{45.0} & 17.5 & 36.9 & \textbf{20.8} & \textbf{38.6}\\
    \bottomrule
  \end{tabular}
\end{table}

\section{Results and Discussion}
\label{sec:discussion}

\textbf{Performance comparison (Q1).}
In Table~\ref{tab:main}, we present the performance of our method. Since predicted boundaries do not align with GT, for each predicted segment $s_j \in \mathcal{S}$, we select the GT annotation with the greatest temporal overlap and treat its label as the unique correct answer. The prediction is counted as correct if the prediction matches such a label. We report top-1 and top-5 accuracy over all predicted segments. As an internal baseline, we run the orchestrator in isolation: it receives each segment's frames and the full action vocabulary $\Lambda$ and directly produces a top-5 ranking, with no specialist deliberation. This isolates the contribution of the specialists protocol from the orchestrator's standalone capability. Following conventional setup used in prior work~\cite{Ragusa_2023_Meccano}, we also report the same evaluation feeding the ground-truth temporal boundaries to the pipeline (GTB columns). To the best of our knowledge, this constitutes the first reported zero-shot evaluation on this dataset. Our proposed method reaches 16.8\% top-1 and 45.0\% top-5 accuracy on average, against 13.5\% and 28.9\% of the baseline. The gain is consistent across all seven test videos with an average increment of $3.1\%$ in top-1 and $16.1\%$ on top-5, indicating that the deliberation provides useful priors to the orchestrator. Under the GTB protocol, our method outperforms the baseline with an average of 20.8\% / 38.6\%  against 17.5\% / 36.9\%.

\textbf{Deliberation influence (Q2).}
A natural concern with a re-ranking step is that the orchestrator may either ignore the agents (yielding gains only by chance) or simply ratify them (in which case its role is secondary). Table~\ref{tab:influence} shows neither is the case. On average, the orchestrator \emph{overrides} its initial top-1 prediction in 70.9\% of segments after deliberation, demonstrating that the specialists' evidence is being utilized. The overrides are net positive: in 21.6\% of segments, the orchestrator flips from an incorrect to a correct top-1 prediction, versus only 10.3\% flips in the opposite direction. The asymmetry is even stronger at top-5, where 28.1\% of segments are re-labeled correctly in the final ranking against only 2.9\% wrongly re-labeled. This is consistent with the protocol design: the Borda aggregation pools complementary labelings from the three heterogeneous specialists, expanding the orchestrator's top-5 reliably while leaving the harder top-1 decision noisier.

\textbf{Heterogeneity ablation (Q3)} Table~\ref{tab:ablation-backbone} replaces the three heterogeneous specialists with three copies of the same backbone, keeping the rest of the pipeline (orchestrator, six-turn protocol, Borda aggregation, re-ranking) unchanged. The three homogeneous configurations reach 15.5/33.5 (Gemma4:e4b), 14.6/42.5 (Ministral-3:8b), and 15.9/43.2 (Qwen3.5:9b) top-1/top-5, respectively, all below the heterogeneous main result of 16.8/45.0. The Gemma4-only configuration, in particular, shows that simply running the protocol with three copies of a weaker backbone yields only marginal gains over its own baseline ($+1.8$/$+4.3$), whereas combining the same three families recovers a substantially larger top-5 improvement. We read this as evidence that 
the benefit of the deliberation protocol is not just from more computational power or "more votes" but from the \emph{decorrelated} priors of the three model families: heterogeneous agents disagree on different cues and the Q\&A round in Stage~2 makes those disagreements actionable.

\textbf{Segmentation quality (Q4).} Table~\ref{tab:seg} reports the zero-shot segmentation produced by the orchestrator. The mean IoU against ground-truth segments is 32.9\%, and the fraction of GT segments covered by at least one predicted segment is 47.5\%. These numbers act as a hard ceiling on our method: any GT action whose temporal extent is not covered by Stage~1 is unrecoverable downstream, regardless of how well the agents deliberate. This partially explains the gap between our method ($16.8$/$45.0$) and GTB accuracies and points to segmentation as the most impactful single component to improve in future work.\looseness=-1

\begin{table}[t]
  \centering
  \setlength{\tabcolsep}{3pt}
  \small

  \begin{minipage}[t]{0.62\linewidth}
    \centering
    \caption{Deliberation effect on orchestrator top-1 / top-5 predictions (\%).}
    \label{tab:influence}
    \begin{tabular}{l r rr rr}
      \toprule
      & & \multicolumn{2}{c}{\textbf{Top-1}}
        & \multicolumn{2}{c}{\textbf{Top-5}} \\
      \cmidrule(lr){3-4}\cmidrule(lr){5-6}
      \textbf{Video} & \textbf{Override}
        & \textbf{Correct} & \textbf{Wrong}
        & \textbf{Correct} & \textbf{Wrong} \\
      \midrule
      0008 & 70.6 & 22.6 & 10.7 & 27.3 & 2.9 \\
      0009 & 72.3 & 17.8 & 12.3 & 33.7 & 1.0 \\
      0010 & 71.4 & 24.1 &  7.3 & 27.6 & 1.6 \\
      0011 & 70.2 & 18.7 &  9.0 & 26.7 & 4.7 \\
      0012 & 70.4 & 18.9 & 10.5 & 30.4 & 3.7 \\
      0019 & 74.2 & 24.6 & 11.0 & 31.4 & 1.9 \\
      0020 & 67.2 & 24.4 & 10.9 & 19.8 & 4.5 \\
      \midrule
      \textbf{Avg} & \textbf{70.9}
                   & \textbf{21.6} & \textbf{10.3}
                   & \textbf{28.1} & \textbf{2.9} \\
      \bottomrule
    \end{tabular}
  \end{minipage}
  \hfill
  \begin{minipage}[t]{0.36\linewidth}
    \centering
    \caption{Stage~1 segmentation quality (\%).}
    \label{tab:seg}
    \setlength{\tabcolsep}{8pt}
    \begin{tabular}{l rr}
    \toprule
    & \multicolumn{2}{c}{\textbf{Qwen3.5:27B}} \\
    \cmidrule(lr){2-3}
    \textbf{Video} & \textbf{mIoU} & \textbf{GT cov.} \\
    \midrule
      0008 & 37.4 & 54.2 \\
      0009 & 36.2 & 40.5 \\
      0010 & 25.8 & 48.4 \\
      0011 & 30.4 & 46.0 \\
      0012 & 38.3 & 57.0 \\
      0019 & 31.0 & 47.0 \\
      0020 & 31.0 & 39.1 \\
      \midrule
      \textbf{Avg} & \textbf{32.9} & \textbf{47.5} \\
      \bottomrule
    \end{tabular}
  \end{minipage}
\end{table}

\begin{table*}[t]
  \centering
  \caption{Homogeneous-agent ablation: top-1/top-5 predictions (\%).}
  \label{tab:ablation-backbone}
  \setlength{\tabcolsep}{4pt}
  \footnotesize
  \begin{tabular}{l rrrr rrrr rrrr}
    \toprule
    & \multicolumn{4}{c}{\textbf{Gemma4:e4b}}
    & \multicolumn{4}{c}{\textbf{Ministral-3:8b}}
    & \multicolumn{4}{c}{\textbf{Qwen3.5:9b}} \\
    \cmidrule(lr){2-5}\cmidrule(lr){6-9}\cmidrule(lr){10-13}
    & \multicolumn{2}{c}{Baseline} & \multicolumn{2}{c}{Multi-agent}
    & \multicolumn{2}{c}{Baseline} & \multicolumn{2}{c}{Multi-agent}
    & \multicolumn{2}{c}{Baseline} & \multicolumn{2}{c}{Multi-agent} \\
    \cmidrule(lr){2-3}\cmidrule(lr){4-5}%
    \cmidrule(lr){6-7}\cmidrule(lr){8-9}%
    \cmidrule(lr){10-11}\cmidrule(lr){12-13}
    \textbf{Video} & T-1 & T-5 & T-1 & T-5
                   & T-1 & T-5 & T-1 & T-5
                   & T-1 & T-5 & T-1 & T-5 \\
    \midrule
    0008 & 11.2 & 31.1 & 15.4 & 34.0 & 12.2 & 30.3 & 13.0 & 40.3 & 11.3 & 30.3 & 14.9 & 45.4 \\
    0009 & 11.9 & 21.8 &  7.9 & 18.8 & 10.9 & 15.8 & 23.8 & 47.5 & 11.8 & 17.6 & 18.0 & 43.6 \\
    0010 & 22.4 & 40.1 & 25.1 & 48.4 & 22.9 & 37.5 & 18.8 & 49.5 & 23.2 & 39.2 & 17.2 & 43.1 \\
    0011 & 13.7 & 35.1 & 15.8 & 34.0 & 18.3 & 39.8 & 11.5 & 48.2 & 17.8 & 36.3 & 15.1 & 51.5 \\
    0012 &  8.9 & 23.0 & 12.3 & 28.1 &  9.6 & 24.4 &  8.9 & 29.6 &  6.7 & 20.7 & 11.4 & 29.2 \\
    0019 & 12.2 & 23.9 & 20.1 & 40.3 & 13.8 & 27.7 & 11.9 & 44.0 & 13.8 & 25.4 & 14.1 & 46.8 \\
    0020 & 15.4 & 29.4 & 12.6 & 31.1 & 17.5 & 28.8 & 14.1 & 38.4 & 18.1 & 29.4 & 20.7 & 43.1 \\
    \midrule
    \textbf{Avg} & 13.7 & 29.2 & \textbf{15.5} & \textbf{33.5} & \textbf{15.0} & 29.2 & 14.6 & \textbf{42.5}  & 14.8 & 28.4 & \textbf{15.9} & \textbf{43.2} \\
    \bottomrule
  \end{tabular}
\end{table*}


\section{Limitations, Conclusion and Future Work}
\label{sec:conclusion}
We have presented \emph{Divide, Deliberate, Decide}, a zero-shot multi-agent framework for fine-grained egocentric action recognition. A VLM orchestrator segments the video and produces an initial top-$k$ ranking, a set of heterogeneous specialist VLMs deliberate over each segment through peer consultation, proposing a refined ranking, fed back to the orchestrator for the final decision. Our experiments indicate that the protocol consistently improves the baseline accuracy of the orchestrator, positively overriding its initial predictions in $\approx 20\%$ of cases. These results suggest that structured deliberation between small, heterogeneous VLMs is a viable alternative to scaling a single model on fine-grained video understanding tasks, while keeping the system on-premise and fine-tuning-free. There are several aspects for future work. First, the quality of the zero-shot action segmentation is the dominant bottleneck of our system. Even a small amount of supervision on action boundaries is likely to provide significantly tighter segments, improving the overall performance of the protocol. A complementary strategy is to recast segmentation as an action anticipation task, asking the orchestrator to predict, conditioned on the past frames, which action is about to occur, and that prediction is used as a prior for where the next boundary should be placed. Finally, agent disagreement could trigger re-segmentation. Windows with low-margin or high-entropy Borda rankings get flagged for a second pass, focusing where it is least confident. Second, the evaluation is restricted to a single benchmark (MECCANO), while the protocol can, in principle, generalize to other egocentric domains. Third, the framework requires eleven VLM calls per segment, which is acceptable for off-line analysis but not yet compatible with real-time deployment. Studying the influence of the number and size of agents on the cost–accuracy trade-off, including asymmetric configurations in which only the most uncertain segments trigger the full deliberation. \looseness=-1

\begin{acknowledgments}
The author gratefully acknowledges the financial support provided by A. Loacker S.p.A. through the industrial scholarship within the PhD Program in Adv. Sys. Eng. (Free University of Bozen-Bolzano).
\end{acknowledgments}

\section*{Declaration on Generative AI}
During the preparation of this work, the author(s) used generative AI to assist with text formatting (Grammarly \& Claude Opus 4.7). All scientific content and analyses were produced by the author(s), who take full responsibility for the final content of the publication.

\bibliography{references}

@misc{NVIDIA_RTX6000,
  author        = "{NVIDIA Corporation}",
  year          = "2026",
  title         = "{NVIDIA RTX 6000 Ada Generation Graphics Card}",
  url           = "https://www.nvidia.com/en-us/products/workstations/rtx-6000/",
  lastaccessed  = "April 27, 2026",
}

@article{Ragusa_2023_Meccano, 
 title={MECCANO: A multimodal egocentric dataset for humans behavior understanding in the industrial-like domain}, 
 journal={CVIU}, 
 publisher={Elsevier BV}, 
 author={Ragusa, Francesco and Furnari, Antonino and Farinella, Giovanni Maria}, 
 year={2023}, 
  }

@inproceedings{Damen_2022_EPIC_KITCHENS,  title={EPIC-KITCHENS VISOR Benchmark: VIdeo Segmentations and Object Relations}, booktitle={Advances in Neural Information Processing Systems 35}, publisher={NeurIPS}, author={Damen, Dima and Darkhalil, Ahmad and Fidler, Sanja and Fouhey, David and Higgins, Richard and Kar, Amlan and Ma, Jian and Shan, Dandan and Zhu, Bin}, year={2022},  collection={NeurIPS 2022} }

@article{Grauman_2025, title={Ego4D: Around the World in 3,600 Hours of Egocentric Video}, volume={47}, number={11}, journal={IEEE Transactions on PAMI}, publisher={IEEE}, author={Grauman, Kristen and Westbury, Andrew and Byrne, Eugene and Cartillier, Vincent and Chavis, Zachary and Furnari}, year={2025} }

@article{yao2022react,
  title={ReAct: Synergizing Reasoning and Acting in Language Models},
  author={Yao, Shunyu and Zhao, Jeffrey and Yu, Dian and Du, Nan and Shafran, Izhak and Narasimhan, Karthik and Cao, Yuan},
  journal={arXiv},
  year={2022}
}

@inbook{Taulli_2025_autogen, title={AutoGen}, ISBN={9798868811340}, booktitle={Building Generative AI Agents}, publisher={Apress}, author={Taulli, Tom and Deshmukh, Gaurav}, year={2025}, pages={147–177} }

@article{li2023camel,
  title={CAMEL: Communicative Agents for Mind Exploration of Large Language Model Society},
  author={Li, Guohao and Hammoud, Hasan and Itani, Hani and Khizbullin, Dmitrii and Ghanem, Bernard},
  journal={ANIPS},
  year={2023}
}

@article{cai2024temporalbench,
  title={Temporalbench: Benchmarking fine-grained temporal understanding for multimodal video models},
  author={Cai, Mu and Tan, Reuben and Zhang, Jianrui and Zou, Bocheng and Zhang, Kai and Yao, Feng and Zhu, Fangrui and Gu, Jing and Zhong, Yiwu and Shang, Yuzhang and others},
  journal={arXiv},
  year={2024}
}

@inproceedings{li2024mvbench,
  title={MVBench: A Comprehensive Multi-modal Video Understanding Benchmark},
  author={Li, Kunchang and Wang, Yali and He, Yinan and Li, Yizhuo and Wang, Yi and Liu, Yi and Wang, Zun and Xu, Jilan and Chen, Guo and Lou, Ping and others},
  booktitle={2024 IEEE/CVF CVPR},

  year={2024}
}

@inproceedings{wang2025inference,
  title={Inference Compute-Optimal Video Vision Language Models},
  author={Wang, Peiqi and Peng, ShengYun and Zhang, Xuewen and Yu, Hanchao and Yang, Yibo and Huang, Lifu and Liu, Fujun and Wang, Qifan},
  booktitle={Proceedings of the 63rd Annual Meeting of the Association for Computational Linguistics (Volume 1: Long Papers)},
  year={2025}
}

@inproceedings{liang2024encouraging,
  title={Encouraging divergent thinking in large language models through multi-agent debate},
  author={Liang, Tian and He, Zhiwei and Jiao, Wenxiang and Wang, Xing and Wang, Yan and Wang, Rui and Yang, Yujiu and Shi, Shuming and Tu, Zhaopeng},
  booktitle={EMNLP},
  year={2024}
}

@article{wang2022self,
  title={Self-consistency improves chain of thought reasoning in language models},
  author={Wang, Xuezhi and Wei, Jason and Schuurmans, Dale and Le, Quoc and Chi, Ed and Narang, Sharan and Chowdhery, Aakanksha and Zhou, Denny},
  journal={arXiv},
  year={2022}
}

@inproceedings{van2000variants,
  title={Variants of the borda count method for combining ranked classifier hypotheses},
  author={Van Erp, Merijn and Schomaker, Lambert},
  booktitle={7th International Workshop on frontiers in handwriting recognition},
  year={2000},
  organization={IUF}
}

@inproceedings{bianchi2026skillformer,
  title={SkillFormer: unified multiview video understanding for proficiency estimation},
  author={Bianchi, Edoardo and Liotta, Antonio},
  booktitle={ICMV 2025},
  year={2026},
  organization={SPIE}
}

@article{bianchi2026profvlm,
  title={ProfVLM: A lightweight video-language model for multi-view proficiency estimation},
  author={Bianchi, Edoardo and Staiano, Jacopo and Liotta, Antonio},
  journal={CVIU},
  year={2026},
  publisher={Elsevier}
}

\end{document}